%% file: main.tex
\documentclass[runningheads]{llncs}

\usepackage{eccv}

\input{preamble}

\input{macros}

\usepackage{eccvabbrv}

\usepackage{graphicx}
\usepackage{booktabs}

\usepackage[accsupp]{axessibility}  

\usepackage{hyperref}

\usepackage{orcidlink}

\begin{document}

\title{We'll Fix it in Post: Improving Text-to-Video Generation with Zero Training} 


\author{Minkyu Choi$^{*\dagger}$\inst{1}\orcidlink{0009-0007-6557-8865} \and
S P Sharan$^\dagger$\inst{1}\orcidlink{0000-0002-6298-6464} \and
Harsh Goel\inst{1}\orcidlink{0009-0006-9873-9584} \and
\\
Sahil Shah\inst{1}\orcidlink{0009-0009-3640-2339} \and
Sandeep Chinchali\inst{1}\orcidlink{0000-0002-0601-3633}}

\authorrunning{M. Choi et al.}

\institute{The University of Texas at Austin TX 78712, USA \\
$^*$Corresponding author: {\tt \small minkyu.choi@utexas.edu}\\
$^\dagger$Contributed equally to this work.}

\maketitle

\begin{abstract}
     Current text-to-video (T2V) generation models are increasingly popular due to their ability to produce coherent videos from textual prompts. However, these models often struggle to generate semantically and temporally consistent videos when dealing with longer, more complex prompts involving multiple objects or sequential events. Additionally, the high computational costs associated with training or fine-tuning make direct improvements impractical. To overcome these limitations, we introduce \(\projectname\), a novel zero-training video refinement pipeline that leverages neuro-symbolic feedback to automatically enhance video generation, achieving superior alignment with the prompts. Our approach first derives the neuro-symbolic feedback by analyzing a formal video representation and pinpoints semantically inconsistent events, objects, and their corresponding failure frames with respect to the prompt. This feedback then guides targeted edits to the original video. Extensive empirical evaluations on both open-source and proprietary T2V models demonstrate that \(\projectname\) significantly enhances temporal and logical alignment across diverse prompts by almost $40\%$.
\end{abstract}

\section{Introduction}
\label{sec:introduction}
    
    Imagine generating the following complex scenario from a text-to-video (T2V) model: 
    \begin{quoting}[leftmargin=0.8cm, rightmargin=0.8cm]
        \textit{``An autonomous vehicle crosses an intersection \uline{after} waiting for a pedestrian to cross."}
    \end{quoting}

    This involves three interdependent aspects: \ding{202} \textit{semantic correctness} (the presence of objects such as the car and the pedestrian, and actions such as the pedestrian walking and the car stopping), \ding{203} \textit{spatial coherence} (entities interact correctly in 3D space, for instance the car stops before the intersection when the pedestrian crosses), and \ding{204} \textit{temporal consistency} (events occur in the correct order over time, for instance, the car stops when the pedestrian crosses and the car moves after the pedestrian is off the road). 

    Current research in T2V generation has primarily focused on the first two aspects. Existing methods improve visual quality and semantic accuracy by modifying rewards in diffusion models, tweaking attention maps, or making low-level architectural changes. Although these enhance visual fidelity, they fail to fix temporal misalignments and are often infeasible for models like Gen3 \cite{gen3} and Pika \cite{pika}, which lack accessible model weights.

    \begin{figure}[t]
        \centering
        \includegraphics[width=\textwidth, trim=0cm 3.39in 7.76in 0cm, clip]{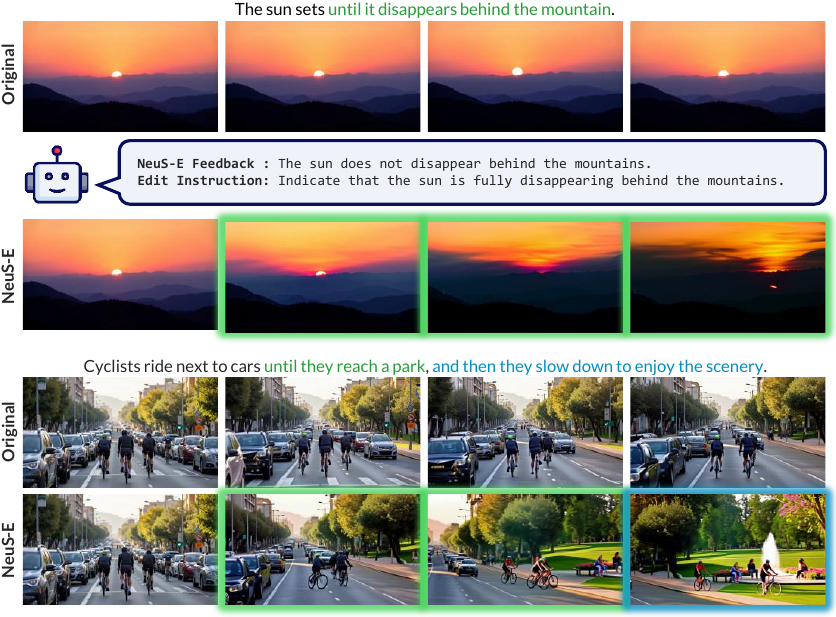}
        
        \caption{\textbf{\(\projectname\) improves the text-to-video (T2V) temporal alignment.} The border color of the frames corresponds to the identified events. Vanilla T2V models (top video) fail to generate the sunset behind the mountains.
        \(\projectname\) systematically identifies and surgically corrects this video segment to improve the temporal fidelity of the synthetic video with targeted feedback.}
        \label{fig:teaser}
    \end{figure}
    
     However, state-of-the-art T2V models fail catastrophically when prompted to generate events in a specific temporal order. Training-free methods that utilize excessive remarking to improve traditional T2V evaluation benchmarks, such as VBench \cite{huang2024vbench},  fail to improve temporal alignment since these metrics prioritize visual aesthetics.  In contrast, \neusv~\cite{sp_2025_CVPR} introduces a neuro-symbolic method that rigorously quantifies a text-to-video model's temporal alignment with respect to the prompt. Taking inspiration from this work, we pose the following research question: 
    
\begin{quoting}[leftmargin=0.50cm, rightmargin=0.50cm] 
    \textit{``Can we leverage neuro-symbolic feedback to surgically refine temporally misaligned video segments through targeted edits, thereby improving text-to-video temporal alignment?"}
\end{quoting}

    We introduce \(\projectname\), a zero-training framework for targeted video refinement. Inspired by \neusv, we devise a neuro-symbolic feedback loop that disentangles atomic events and objects (termed propositions) with their prompted temporal order (termed specifications). Therefore, instead of treating videos as static outputs, \(\projectname\) identifies video segments that weakly satisfy the objects and events in the prompt, edits the corresponding keyframes, and regenerates only these misaligned portions to improve temporal alignment with respect to the prompt. This iterative process produces a video that aligns with the prompt's temporal requirements without retraining the generative model. 

    A key insight behind our approach is that improving T2V generation does not \textit{necessarily} require modifying the generative model itself. Instead, we treat generation errors as diagnosable failures of a temporal specification. By formally verifying the generated video against a temporal logic representation of the prompt, we can identify which proposition is violated and where the failure occurs in time. This diagnostic signal allows us to target only the portion of the video responsible for the violation, enabling localized refinement rather than repeated full-video regeneration. Our key contributions are summarized as follows:
    

    \begin{itemize}[leftmargin=2ex]
        \item We develop a method to extract neuro-symbolic feedback, identifying weakly satisfied propositions and their corresponding problematic video segments.
        \item We introduce \(\projectname\), which automatically corrects weak segments through neuro-symbolic feedback by identifying and editing the keyframe, then regenerating misaligned video segments to improve temporal alignment.

        \item  We demonstrate that \(\projectname\) significantly boosts the temporal fidelity of both open-source (CogVideoX) and closed-source (Gen-3, Pika) models, all with zero additional training.
    \end{itemize}

\section{Related Work}
\label{sec:related_work}
    \paragraph{Text-to-Video Generation.}
        
        Text-to-video models such as SORA \cite{openai2024sora}, GEN-3 Alpha \cite{runway2024gen3}, Kling \cite{kling}, Veo \cite{veo2024}, Wan \cite{wan2.1}, and PIKA \cite{pikaai2024} have seen significant advancements.
        The underlying research follows two main training strategies. The first involves end-to-end training of spatial and temporal modules using diffusion \cite{ho2020denoising} or autoregressive \cite{hong2022cogvideo, yang2024cogvideox} architectures, encompassing early latent-space works \cite{villegas2022phenaki, zhang2023i2vgen, esser2023structure} and numerous modern systems \cite{chen2023videocrafter1opendiffusionmodels, chen2024videocrafter2, xing2024make, zhang2024moonshot, zhang2024show, Ho2022VideoDM, he2023latentvideodiffusionmodels, kong2025hunyuanvideosystematicframeworklarge, wang2024videofactory, wan2.1}. The second strategy efficiently adapts pre-trained image models by training only a lightweight temporal module \cite{blattmann2023align, Guo2023AnimateDiffAY, khachatryan2023text2videozerotexttoimagediffusionmodels}. For a broader survey, see \cite{cho2024sora}.
        In contrast to these methods, \projectname\ proposes a training-free approach to video generation, for temporally complex prompts. 
        
    \paragraph{Text-to-Video Refinement.}

        Training-free frameworks improve text-to-video generation by leveraging text-to-image (T2I) models \cite{Meng2021SDEditGI, Brooks2022InstructPix2PixLT, zhang2023adding} to edit video frames \cite{Jeong2023GroundAVideoZG, Zhang2024MoonshotTC, khachatryan2023text2videozerotexttoimagediffusionmodels, Yang2023RerenderAV, wang2024zeroshotvideoeditingusing, goel2024improving, goel2024syndiff}. Most of these methods focus on improving the temporal coherence of individual objects by rectifying cross-frame attention maps and features \cite{Geyer2023TokenFlowCD, Jeong2023GroundAVideoZG, Qi2023FateZeroFA, Yang2023RerenderAV, Liu2023VideoP2PVE}, by refining text prompts \cite{kim2024free2guidegradientfreepathintegral, luo2025enhanceavideobettergeneratedvideo}, by editing keyframes \cite{Ceylan2023Pix2VideoVE, zhang2024moonshot}, or employing video-to-video diffusion models \cite{molad2023dreamixvideodiffusionmodels}. While effective for object-level consistency, these approaches fail to enforce the logical order of multiple, distinct events (e.g., ensuring a person enters a car before driving away). To bridge this critical gap, \(\projectname\) uses neuro-symbolic feedback to guide edits, ensuring the final video satisfies complex temporal relationships across the entire event sequence.

    \paragraph{Evaluation.}

    Current Text-to-Video (T2V) evaluation methods include distribution-based metrics (FID, FVD, CLIPSIM) \cite{shin2024lostmelodyempiricalobservations, liu2023fetvbenchmarkfinegrainedevaluation, jain2024peekaboointeractivevideogeneration, bugliarello2023storybenchmultifacetedbenchmarkcontinuous, hu2022makemovecontrollableimagetovideo, yu2023celebvtextlargescalefacialtextvideo}, LLM-based VQA scoring \cite{kou2024subjectivealigneddatasetmetrictexttovideo, liu2024evalcrafterbenchmarkingevaluatinglarge, zhang2024benchmarkingaigcvideoquality, Li_2024_CVPR}, and metric ensembles for visual and spatial quality like VBench, EvalCrafter, FETV, and T2V-Bench \cite{huang2024vbench, liu2024evalcrafter, liu2024fetv, ji2024t2vbench, feng2024tc, huang2023t2i, chu2024sora}. While comprehensive, these approaches primarily assess aesthetic temporal quality, such as motion and flickering \cite{huang2024vbench}, rather than logical event sequencing. In contrast, \(\projectname\) leverages \neusv~\cite{sp_2025_CVPR} to address this gap, using Temporal Logic (TL) to formally verify complex temporal relationships.

    \paragraph{Formal Verification.}

    Formal verification methods are used to construct symbolic representations of events and tasks across various domains. These methods have been applied in video understanding \cite{short1, short2, event-detection-video-stream, cnn-event-detect, lstm-event-detect, neural-symbolic, chen2022comphy}, robotics \cite{shoukry2017linear, hasanbeig2019reinforcement, kress2009temporal}, and autonomous driving \cite{jha2018safe, mehdipour2023formal}, using techniques like graph-based reasoning \cite{graphical-model-relationship-detection, visual-symbolic, long3}, latent-space abstractions \cite{symbolic-high-speed-video, BertasiusWT21, neural-symbolic-cv}, or formal languages \cite{Baier2008}. In contrast, \(\projectname\) leverages formal verification to provide structured feedback—identifying text-based edit instructions and keyframes—to improve Text-to-Video generation without any training.


\section{Preliminaries}
\label{sec:preliminaries}

    \begin{figure}[t]
        \centering
        \includegraphics[width=1.0\linewidth]{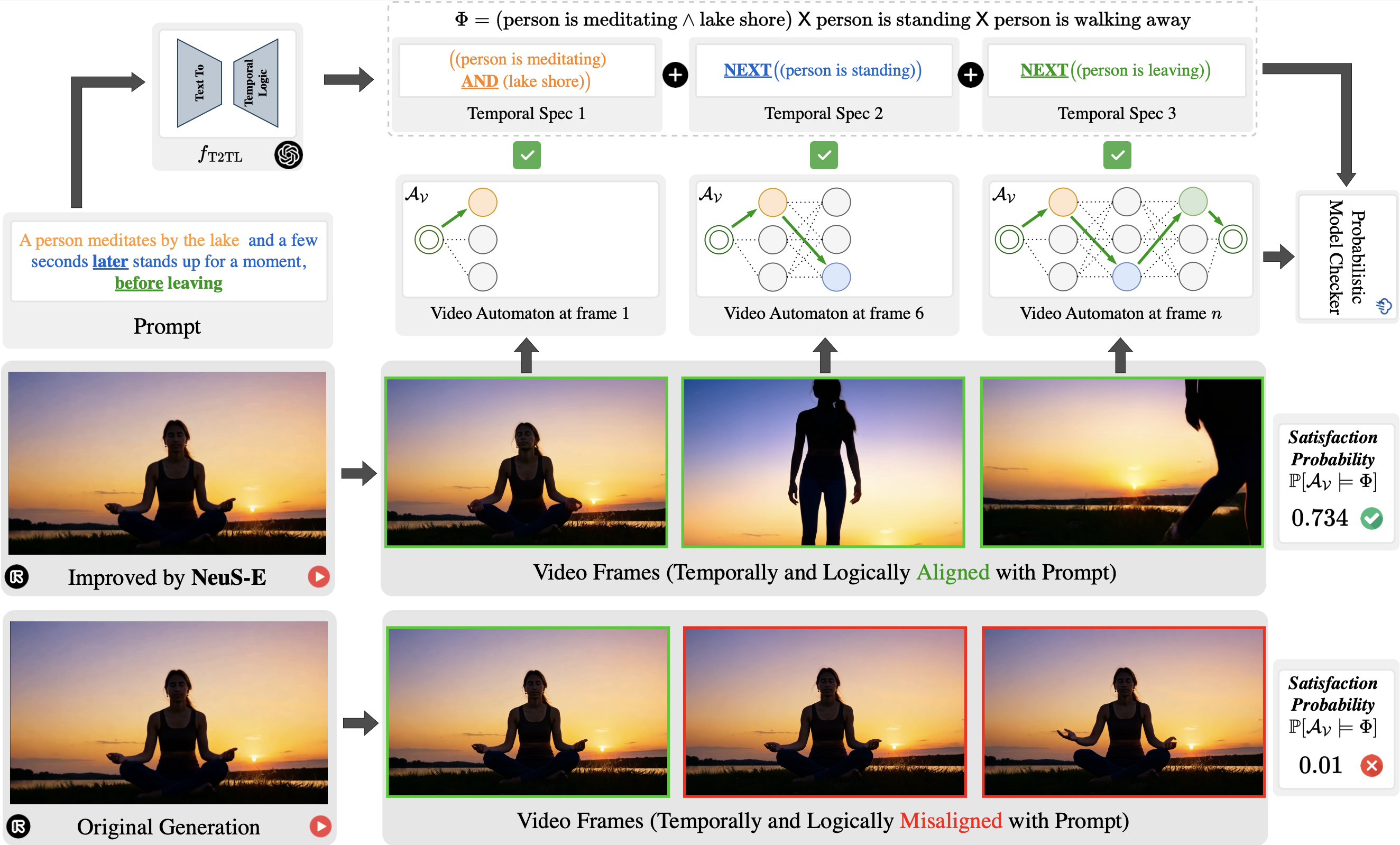}
        \caption{\textbf{Formally verify generated video with video Automaton}. The video automaton expands as new frames are added. Once fully constructed, we verify it against the TL specification. We have a very low probability of satisfaction from the initial generation, as the person neither stands nor walks away. To address this, we refine the video using \(\projectname\), generating a better video that is temporally and logically aligned with the prompt and achieves a higher satisfaction probability.}
        \label{fig:video_automaton_example}
    \end{figure}
    
    We begin with a comprehensive running example of generating a promotional video using the prompt: ``A person meditates by the lake and, a few seconds later, stands up for a moment before leaving" to describe the following terminologies. 
    
    
    \paragraph{Temporal Logic.}


        Temporal Logic (TL) is an expressive formal language that combines logical and temporal operators to express time-dependent statements \cite{Temporal-and-Modal-Logic, Manna}. A TL specification or formula is structured around three key components: \ding{202} a set of atomic propositions, \ding{203} first-order logic operators, and \ding{204} temporal operators. Atomic propositions are fundamental statements that evaluate to either \texttt{True} or \texttt{False} and serve as building blocks for more complex expressions. First-order logic operators include AND ($\wedge$), OR ($\vee$), NOT ($\neg$), IMPLY ($\Rightarrow$)., and the temporal operators consist of ALWAYS (\always), EVENTUALLY (\eventually), NEXT (\ournext), UNTIL (\until), etc. The set of atomic propositions $\propset$, and TL specification $\spec$ of our running example are:
\begin{equation}
    \label{eq:eg_spec}
    \begin{array}{rl}
        \propset &= \{\text{person is meditating}, \ \text{lake shore}, 
        \text{person is standing}, \text{person is walking away}\}. \\ 
        \Phi &= (\text{person is meditating} \wedge \text{lake shore}) \mathsf{X}
        (\text{person is standing} \ \mathsf{X} \ \text{person is walking away}).
    \end{array}
\end{equation}

    \paragraph{Video Automaton.}
        A video automaton formally represents a video sequence as a sequence of states and transitions \cite{choi2025towards, sp_2025_CVPR, yang2023specification}. It is a form of Discrete-Time Markov Chain (DTMC) \cite{norris1998markov, kemeny1960finite} used to model generated videos as each frame's transition depends only on the current frame. Since video sequences are inherently \textit{discrete}, \textit{finite}, and \textit{temporal}, DTMCs can approximate frame transitions. To this end, we define the set of states in the video as $Q$ and this frame transition function as $\delta:Q \times Q \to [0,1]$ where, given two states $q_1 \in Q$ and $q_2 \in Q$, $\delta(q_1, q_2) \in [0,1]$ gives the probability of transitioning from $q_1$ to $q_2$. Now, the video automaton, $\va$, is the tuple $\va=(Q,q_0,\delta,\lambda)$ where the initial state is $q_0 \in Q$ and the label function is $\lambda:Q \to 2^{|\propset]}$.

    \paragraph{Formal Verification.}
        Formal verification ensures formal guarantees that the system satisfies the desired specification. \cite{clarke1999model, huth2004logic}. It necessitates a formal representation of the system, such as a finite-state automaton (FSA). Given the video automaton $\va$, a \textit{path} in a video is defined as a sequence of states $q_0 q_1 (q_2)^\omega$ starting from the initial state $q_0$ and $\omega$ denotes repetition.
        A trace corresponds to the sequence of labels $\lambda(q_0) \lambda(q_1) \lambda(q_2) \cdots \in (2^{|\propset|})^\omega$ associated with the states along a path, where $\lambda(q)$ represents the labeling function that maps each state $q$ to a subset of atomic propositions from the set $\propset$. The trace captures the progression of events or properties over time as observed along the path. Next, we apply probabilistic model checking~\cite{Baier2008} to determine how much the \textit{trace} starting from the initial state satisfies the temporal logic (TL) specification $\Phi$. Using these formal representations, we evaluate the videos.
    
    \paragraph{Text to Temporal Logic.}
    Text prompts are converted to temporal logic conversion using LLMs \cite{chen2024nl2tltransformingnaturallanguages, cosler2023nl2specinteractivelytranslatingunstructured, Mendoza2024SYNTHTL, yang2023specification, sp_2025_CVPR} for subsequent analysis. We denote the LLM-based text-to-temporal logic (T2TL) function as $f_{\text{T2TL}} : T \to (\propset, \spec)$ to decompose a text prompt $T$ into a TL specification $\spec$ and a set of propositions $\propset$. A prompt example is presented in the Appendix.
    
    \paragraph{Neuro-symbolic Video Verification.} We uniquely adapt a VLM to construct the video automaton and formally verify it as a symbolic method. We calibrate the VLM model to map its naive confidence to the accuracy. We present a detailed explanation of the VLM calibration process and usage in the Appendix.
    \begin{definition}[Video Automaton Construction]  
    \label{def:video_automaton_construction}
        Given a generated video $\video$ and a set of atomic propositions $\propset$, we construct a video automaton. For each proposition \(\prop_i \in \propset\) and frame \(\frame_n \in \video\), a VLM computes a semantic confidence score: \(\vlm: \prop_i \times \frame_n \to c_{i,n}\), where \(c_{i,n} \in [0,1]\) represents the confidence of \(\prop_i\) in \(\frame_n\). For each frame $n$, we define the confidence set as: \(\mathbb{C} = \{ c_{i,n} \mid \prop_i \in \propset, \frame_n \in \video\ \}\). The video automaton $\va$ is then generated by applying a function $\xi$ that processes the set of propositions $\propset$ along with the confidence scores across all frames:
            \begin{equation} 
                \label{eq:va}
                \xi : \propset \times \mathbb{C} \to \va.
            \end{equation}
    \end{definition}

    \begin{definition}[Satisfaction Probability] 
    \label{def:satisfaction_probability}
        Given a video automaton \(\va\) and a temporal logic specification \(\spec\), the satisfaction probability \(\probsatis\) is computed by verifying \(\va\) against \(\spec\) using STORM \cite{storm, stormpy}, which uses probabilistic computation tree logic (PCTL). This verification process is formalized as a probabilistic model checking function:  
        \begin{equation}  
            \label{eq:satisfaction}
             \Psi : \va \times \spec \to \probsatis.
        \end{equation}   
    \end{definition}  
    We provide details on the video automaton generation function in the Appendix.


\section{Methodology}
\label{sec:methodology}


    
    

Given a synthetic video $\mathcal{V}$ generated by T2V models $\mathcal{M_\text{T2V}}: T , I \to \mathcal{V}$, where $T$ represents the input text prompt, and $I$ is the \textit{optional} input image, our objective is to refine $\mathcal{V}$ to improve its temporal and logical consistency with the intended semantics of $T$. We introduce \projectname, a neuro-symbolic framework that operates through the following iterative steps:

\begin{itemize}
    \item \textbf{Step 1: Decompose \& Represent.} The prompt $T$ is decomposed into a TL specification, and a video automaton is constructed from $\mathcal{V}$ using a VLM.

    \item \textbf{Step 2: Identify Errors.} By analyzing the video automaton, the framework identifies weak propositions and their corresponding frames, pinpointing temporal misalignments.

    \item \textbf{Step 3: Refine \& Iterate.} The identified errors are converted into keyframe editing instructions to guide a video editing pipeline. This process is repeated until the refined video meets a predefined coherence threshold.
\end{itemize}


    \subsection{Neuro-symbolic Video Verification.}
        First, we generate the video with the prompt without a input image: $\mathcal{V} = \mathcal{M_\text{T2V}}(T,\text{None})$. Then, we decompose $T$ into the TL specification $\spec$ and the set of propositions $\propset$. Given $\spec$ and $\propset$, we construct the video automaton as described in \Cref{def:video_automaton_construction} and verify it according to \Cref{def:satisfaction_probability}. Finally, we perform a comprehensive neuro-symbolic verification of the generated video. Further details are provided in \Cref{fig:video_automaton_example}.

    \subsection{Identifying the Weakest Proposition.}

    In our running example (see \Cref{fig:video_automaton_example}), the generated video fails to satisfy the full Temporal Logic (TL) specification---$(\text{A} \wedge \text{B}) \ \mathsf{X} \ \text{C} \ \mathsf{X} \ \text{D}$---because it omits the events corresponding to propositions C and D. To pinpoint which proposition is the weakest link, we systematically measure the impact of each one on the video's overall satisfaction probability, $\probsatis$.

Our method creates a hypothetical scenario for each proposition $\ep \in \propset$ where we assume it is perfectly satisfied. Intuitively, we assess the video against partial specifications, such as ($B$) $\mathsf{X}$ $C$ $\mathsf{X}$ $D$, ($A$) $\mathsf{X}$ $C$ $\mathsf{X}$ $D$, ($A$ $\wedge$ $B$) $\mathsf{X}$ $D$, and ($A$ $\wedge$ $B$) $\mathsf{X}$ $C$, where each proposition's confidence score is systematically adjusted to find the weakest proposition.  We achieve this by generating an adjusted confidence score set $\tilde{\mathbb{C}}$ that forces the confidence of the proposition under evaluation, $\ep$, to 1.0, while all other scores remain unchanged:
\begin{equation}
    \label{eq:weak_prop_conf_set}
    \tilde{\mathbb{C}} = \{ \tilde{c}_{j,n} : \tilde{c}_{j,n} = 1.0 \text{ if } p_j = \ep; \quad \tilde{c}_{j,n} = c_{j,n} \text{ if } p_j \neq \ep, \; \forall p_j \in \propset \}.
\end{equation}

Using this adjusted set, we construct a new evaluating video automaton $\evai = \xi(\propset, \tilde{\mathbb{C}})$ and compute a new satisfaction probability $\eprobsatis = \Psi(\evai, \spec)$. The proposition whose forced satisfaction causes the largest increase in this probability is identified as the weakest. We formalize this by finding the proposition $\wp$ that maximizes the difference, $\delta_i$:
\begin{equation}
    \label{eq:weakprop}
    \wp = \arg\max_{i \in |\propset|} \{\delta_i \ : \ \delta_i = \eprobsatis - \probsatis\}.
\end{equation}
The proposition $\wp$ with the highest $\delta_i$ value is the component most responsible for the video's failure to align with the prompt, allowing us to target it for refinement.

    \subsection{Localizing the Influence of the Weakest Proposition.}
    Given the weakest proposition $\wp$, we now localize its influence across $\frame_n \in \video$ with respect to $\wp$. In our running example, after we identify the weakest proposition (\textit{e.g.,} ``person is standing'') from the original generation, we determine the most impacted frames by the proposition, specifically the middle frame in the \Cref{fig:video_automaton_example}. 
    
    For each frame $\frame_n$, we construct the evaluating automaton $\evan$ for the video segment from the initial frame upto $\frame_n$. We construct a new confidence measure $\hat{c}_{i,n}=1.0$ for the weakest proposition $\wp$ at $\frame_n$. For all other frames $\{\frame_m: m \in [0, n-1] \}$ and propositions $p_i \in \propset$ including $\wp$, we set $\hat{c}_{i,n} = c_{i,n} + \gamma$, where $\gamma$ introduces controlled noise to enhance numerical stability in the model-checking process. Subsequently,  we define the set of per-frame satisfaction probabilities $\mathbb{Z}=\{\enprobsatis = \Psi(\evan, \spec)\ \mid \frame_n \in \video\}$. Finally, we obtain the most impacted frame $\frame^{*}_n$ by $\ep$ as follows:
    \begin{equation}
        \label{eq:impactedframe}
            \frame^{*}_n = \arg\max_{n \in \{1, 2, \ldots, N\}} \{z_n \in \mathbb{Z} \mid z_n = \enprobsatis\},
    \end{equation}
    where $N$ is the total number of frames in $\video$, and $\frame^{*}_n$ denotes the frame at which enforcing $\wp$'s certainty yields the greatest satisfaction probability, thus identifying the critical time-step most sensitive to $\wp$'s variability.
    
    \subsection{Video Refinement Using Neuro-Symbolic Feedback}
    We refine the video iteratively to enhance the temporal consistency of the generated video $\mathcal{V}$ through the above-mentioned neuro-symbolic feedback, which includes the baseline satisfaction probability $\probsatis$, the weakest proposition $\wp$, and the most impacted frame $\frame^{*}_n$, 

    \paragraph{Video trimming.} We index the frame $\frame^{*}_n$ in video $\mathcal{V}$ and trim the video to this index to obtain a segment to $\frame^{*}_n$ as: \(\mathcal{V}_{\text{trimmed}} = \mathcal{V}[0:n^{*}]\), where $n^{*}$ is the index of $\frame^{*}_n$. This isolates the portion of the video most affected by $\wp$, preparing it for targeted regeneration.

    \paragraph{New Video Segment Generation.} We use an LLM to generate a new prompt \( T_{\text{new}} \) to generate the next video segment after \( \mathcal{V}_{\text{trimmed}} \) by providing the weak proposition \( \wp \) along with the original prompt \( T_{\text{video}} \). We present all prompts in the Appendix.


    \paragraph{Iterative Video Generation and Edition.} First, we merge the new video segment $\mathcal{V}_{\text{new}}$ with $\mathcal{V}_{\text{trimmed}}$, and repeat the neuro-symbolic feedback process as above to obtain the weakest proposition and the most impacted frame. This process is iterated until $\probsatis$ surpasses a predefined threshold or the maximum iteration limit is reached. We present the algorithm in the Appendix.

\section{Experimental Setup}
\label{sec:exp}
    Following \neusv, we use GPT-4o \cite{openai2023gpt4} to decompose prompts into constituent propositions and generate temporal logic specifications. InternVL2.5-8B \cite{chen2024expanding} serves as the VLM for obtaining per-frame confidence scores during automata construction. For further details on this process, we refer readers to \neusv~\cite{sp_2025_CVPR}.
    
    Once our method identifies the weakest proposition and the most impacted frame, we use GPT-4o to generate the video continuation prompt.
    Then, the video is regenerated using the continuation prompt.
    This process iterates until either (1) the \neusv~score surpasses 0.7, or (2) the number of iterations matches the number of propositions---ensuring that each weak proposition undergoes at least one round of refinement. Finally, we use MoviePy \cite{MoviePy} for video trimming and stitching to integrate the edited video clips into a final MP4 (see \Cref{fig:teaser}).

    \paragraph{T2V Models.}
        We evaluate our method on both closed-source and open-source T2V models to demonstrate its broad applicability. Specifically, we conduct experiments on Gen-3 \cite{runway2024gen3} by RunwayML and Pika-2.2 \cite{pikaai2024} by PikaArt as representatives of closed-source models, and CogVideoX-5B \cite{yang2024cogvideox} as an open-source model. While our approach is designed to be \textit{model-agnostic} and can be applied to \textit{any T2V system}, we select these models due to their widespread use in the community and their ability to accept image inputs. 
    \paragraph{Dataset and Metrics.}
        We use the \neusv \cite{sp_2025_CVPR} prompt suite, which contains temporally extended prompts that challenge T2V models. It spans four themes (Nature, Human \& Animal Activities, Object Interactions, Driving Data) and three complexity levels (Basic, Intermediate, Advanced). We evaluate both original and edited videos using the \neusv~score to measure temporal fidelity, and VBench\cite{huang2024vbench}, a widely used metric for synthetic video quality.

\section{Results}
\label{sec:results}
    Building on our experimental setup, we now evaluate \(\projectname\) through empirical analysis. Our experiments aim to answer three key questions that highlight the necessity of our approach:

    \begin{enumerate}[leftmargin=2em]
        \item Does \(\projectname\) boost \textit{alignment} without sacrificing \textit{aesthetics} on temporally extended prompts?
        \item How does iterative refinement through neuro-symbolic feedback compare to a sequential generation, where complex prompts are carefully broken down and generated sequentially?
        \item Does each iteration meaningfully refine the video, progressively improving temporal fidelity?
    \end{enumerate}


    

    
    \subsection{Improving Text-to-Video Generation with Neuro-Symbolic Feedback}
        As outlined earlier, we benchmark state-of-the-art T2V models, including Gen-3 and Pika as closed-source models, and CogVideoX-5B \cite{yang2024cogvideox} as an open-source alternative.

        \paragraph{Benchmarking \(\projectname\) Refined Videos.}
            We evaluate the before-and-after performance on the \neusv~prompt suite in \Cref{tab:neuse-results}, measuring improvements across different themes and complexity levels. By structuring results this way, we gain deeper insights into where and how \(\projectname\) enhances temporal alignment. At a glance, \(\projectname\) consistently improves text-to-video alignment across all themes and complexities. Notably, the improvement is more pronounced for higher-complexity prompts, where regular generation with T2V models struggle the most. This suggests that while baseline models perform poorly on complex temporal relationships, our refinement method helps retain high temporal fidelity even as prompt difficulty increases. Additionally, we observe that Pika-2.2 excels at following video editing instructions, achieving an over 40\% improvement in \neusv~scores—outperforming other models in leveraging keyframe-based refinements.

            \begin{table*}[t]
                \centering
                \caption{\textbf{\neusv~Score Improvements.} Comparison of original and edited \neusv~scores across different themes and complexity levels for Gen3, Pika, and CogVideoX.}
                \vspace{-1em}
                \resizebox{\linewidth}{!}{
                    \begin{tabular}{cccccccc}
                        \toprule
                        \multicolumn{2}{c}{\multirow{2}{*}{Prompts}} & \multicolumn{2}{c}{Gen-3} & \multicolumn{2}{c}{Pika-2.2} & \multicolumn{2}{c}{CogVideoX-5B} \\
                        \cmidrule(lr){3-4} \cmidrule(lr){5-6} \cmidrule(lr){7-8}
                        &  & Original & Edited & Original & Edited & Original & Edited \\
                        \midrule
                        \multirow{4}{*}{By Theme} 
                            & Nature & 0.581 & 0.677 \textcolor{teal}{(+0.096)} & 0.579 & 0.856 \textcolor{teal}{(+0.277)} & 0.481 & 0.623 \textcolor{teal}{(+0.142)} \\
                            & Human \& Animal Activities & 0.613 & 0.767 \textcolor{teal}{(+0.153)} & 0.638 & 0.872 \textcolor{teal}{(+0.235)} & 0.493 & 0.596 \textcolor{teal}{(+0.103)} \\
                            & Object Interactions & 0.610 & 0.721 \textcolor{teal}{(+0.111)} & 0.420 & 0.707 \textcolor{teal}{(+0.287)} & 0.454 & 0.610 \textcolor{teal}{(+0.156)} \\
                            & Driving Data & 0.546 & 0.611 \textcolor{teal}{(+0.065)} & 0.676 & 0.810 \textcolor{teal}{(+0.134)} & 0.565 & 0.681 \textcolor{teal}{(+0.116)} \\
                        \midrule
                        \multirow{3}{*}{By Complexity} 
                            & Basic (1 TL op.) & 0.723 & 0.781 \textcolor{teal}{(+0.059)} & 0.694 & 0.840 \textcolor{teal}{(+0.146)} & 0.621 & 0.738 \textcolor{teal}{(+0.117)} \\
                            & Intermediate (2 TL ops.) & 0.480 & 0.633 \textcolor{teal}{(+0.153)} & 0.480 & 0.795 \textcolor{teal}{(+0.315)} & 0.387 & 0.540 \textcolor{teal}{(+0.153)} \\
                            & Advanced (3 TL ops.) & 0.370 & 0.527 \textcolor{teal}{(+0.157)} & 0.373 & 0.729 \textcolor{teal}{(+0.356)} & 0.344 & 0.449 \textcolor{teal}{(+0.105)} \\
                        \midrule
                        \multicolumn{2}{c}{Overall Score} & 0.587 & 0.694 \textcolor{teal}{(+0.107)} & 0.577 & 0.811 \textcolor{teal}{(+0.233)} & 0.499 & 0.628 \textcolor{teal}{(+0.129)} \\
                        \bottomrule
                    \end{tabular}
                }
                \vspace{-1em}
                \label{tab:neuse-results}
            \end{table*}

        \paragraph{Do humans agree?}

\begin{figure}[t]
                \centering
                \includegraphics[width=0.8\columnwidth]{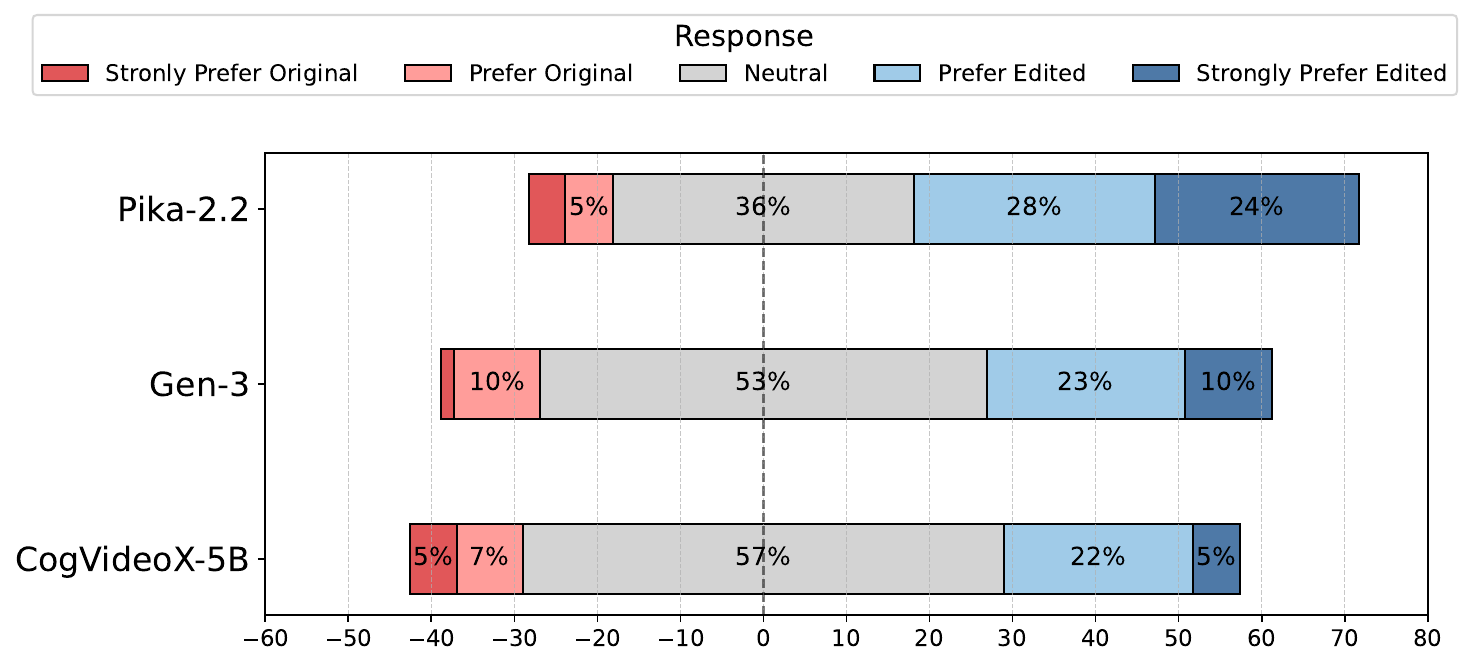}
                \caption{\textbf{Human Evaluation on Video Editing.} Diverging bar chart of human preference labels on the dataset shows that our editing pipeline improves temporal fidelity.}
                \label{fig:human-eval}
            \end{figure}

            To further validate the effectiveness of our approach, we conducted a fully-blind, randomized A/B human evaluation. For each prompt, annotators were shown two videos (original vs. edited) in shuffled order and asked to judge which one better aligned with the caption on a five-point scale: Strongly Agree, Agree, Neutral, Disagree, Strongly Disagree. As shown in \Cref{fig:human-eval}, our edits are preferred in 52\% of trials, closely mirroring the trend observed in our benchmark evaluations. Notably, Pika-2.2 shows the largest improvement, with nearly half of its videos rated as better after refinement. Other models also exhibit consistent gains, and importantly, the number of videos rated as worse remains low across all models. A substantial portion of neutral responses typically arises from (1) videos that were already well-aligned, making edits unnecessary, or (2) cases where T2V models failed to reliably follow regeneration instructions despite multiple iterations. The latter further highlights an inherent limitation of current T2V models in generating specific complex instructions.

    \paragraph{External Benchmark Validation.}
    To further validate that our improvements are not circular with respect to the \neusv~metric, 
    we evaluate \(\projectname\) on T2VCompBench~\cite{sun2025t2vcompbench}, an independent 
    compositional benchmark. On average, \(\projectname\) yields a $+11\%$ improvement across 
    the seven T2VCompBench dimensions, driven primarily by substantial gains in \texttt{Action} 
    and \texttt{Interaction}, which are most closely tied to correcting missing or misordered 
    events. We include category-wise improvements in \Cref{tab:t2vcompbench}.

    \begin{table}[t]
        \centering
        \caption{\textbf{T2VCompBench Scores Before and After Editing.} Category-wise improvement trends for Pika-2.2 under \(\projectname\) refinement.}
        \resizebox{0.55\linewidth}{!}{
            \begin{tabular}{ccc}
                \toprule
                T2VCompBench Category & Original & Edited \\
                \midrule
                Consist-Attr & 0.590 & 0.700 \textcolor{teal}{(+0.110)} \\
                Dynamic-Attr & 0.055 & 0.155 \textcolor{teal}{(+0.100)} \\
                Spatial & 0.510 & 0.600 \textcolor{teal}{(+0.090)} \\
                Motion & 0.270 & 0.390 \textcolor{teal}{(+0.120)} \\
                Action & 0.510 & 0.660 \textcolor{teal}{(+0.150)} \\
                Interaction & 0.580 & 0.710 \textcolor{teal}{(+0.130)} \\
                Numeracy & 0.240 & 0.320 \textcolor{teal}{(+0.080)} \\
                \midrule
                \textbf{Avg} & 0.394 & 0.505 \textcolor{teal}{(+0.111)} \\
                \bottomrule
            \end{tabular}
        }
        \label{tab:t2vcompbench}
        \vspace{-1em}
    \end{table}


    \subsection{On Refinement Strategy -- Is Neuro-Symbolic Feedback Key?}

        One could argue that our method is simply re-prompting a T2V model until the desired output is achieved. Conceptually, the \(\projectname\) pipeline contains two separable components: (1) diagnostic localization, which identifies the weakest proposition in the temporal logic specification and determines the frame where the failure occurs, and (2) execution, which regenerates the video segment conditioned on that diagnosis. Prompt rewriting occurs only in the second step. The key contribution of NeuS-E lies in the first step—the ability to formally diagnose \textit{where} and \textit{why} a video violates the temporal specification. To isolate this effect, we compare this effect with a baseline that performs sequential generation \uline{without} this diagnostic signal (step-by-step generation). 
        Our results (\Cref{tab:ablation-generation-strategy}), this baseline provides only marginal improvements, indicating that the gains arise primarily from neuro-symbolic localization rather than repeated prompting alone.
        
        
        The key advantage of \(\projectname\) lies in its ability to extract two critical insights: (1) identifying weak propositions in the video and (2) determining the optimal time for those propositions to occur to best satisfy the temporal logic (TL) specification. Formal verification adds rigor to our approach. Additionally, we observe that step-by-step generation leads to longer videos, suggesting that without targeted neuro-symbolic feedback, the model produces redundant content rather than precisely correcting temporal inconsistencies.

        A key strength of generating the full video first is that the T2V model establishes a unified global context across the entire timeline. In contrast, step-by-step generation treats segments in isolation or as auto-regressive extensions, which frequently leads to semantic drift (\eg, an actor's appearance changing or the background shifting between steps). Furthermore, in the sequential approach, a minor hallucination in step $t$ becomes the ground truth for step $t{+}1$; these errors accumulate and compound over the sequence. By generating the full video first and then applying surgical edits, \(\projectname\) retains the ``strong'' segments that benefit from global context and only subjects the ``weak'' segments to targeted, simplified regeneration.

        \begin{table}[t]
            \centering
            \caption{\textbf{Ablation Study on Refinement Strategies.} Comparison of \neusv~and VBench scores for Pika-2.2 using neuro-symbolic feedback versus step-by-step prompting. Neuro-symbolic feedback is key for video refinement.}
            \resizebox{0.8\linewidth}{!}{
                \begin{tabular}{cccccc}
                    \toprule
                    \multirow{2}{*}{Strategy} & \multicolumn{2}{c}{\neusv} & \multicolumn{2}{c}{VBench} & \multirow{2}{*}{Length} \\
                    \cmidrule(lr){2-3} \cmidrule(lr){4-5}
                    & Original & Edited & Original & Edited \\
                    \midrule
                    Neuro-Symbolic & 0.577 & 0.811 \textcolor{teal}{($+$0.233)} & 0.789 & 0.772 \textcolor{red}{($-$0.017)} & 11.2 sec \\
                    Step-by-Step & 0.577 & 0.612 \textcolor{teal}{($+$0.035)} & 0.789 & 0.784 \textcolor{teal}{($+$0.005)} & 14.7 sec \\
                    \bottomrule
                \end{tabular}
            }
            \label{tab:ablation-generation-strategy}
            \vspace{-1em}
        \end{table}

        \begin{figure}[t]
            \centering
            \begin{subfigure}{0.32\linewidth}
                \centering
                \includegraphics[width=\linewidth]{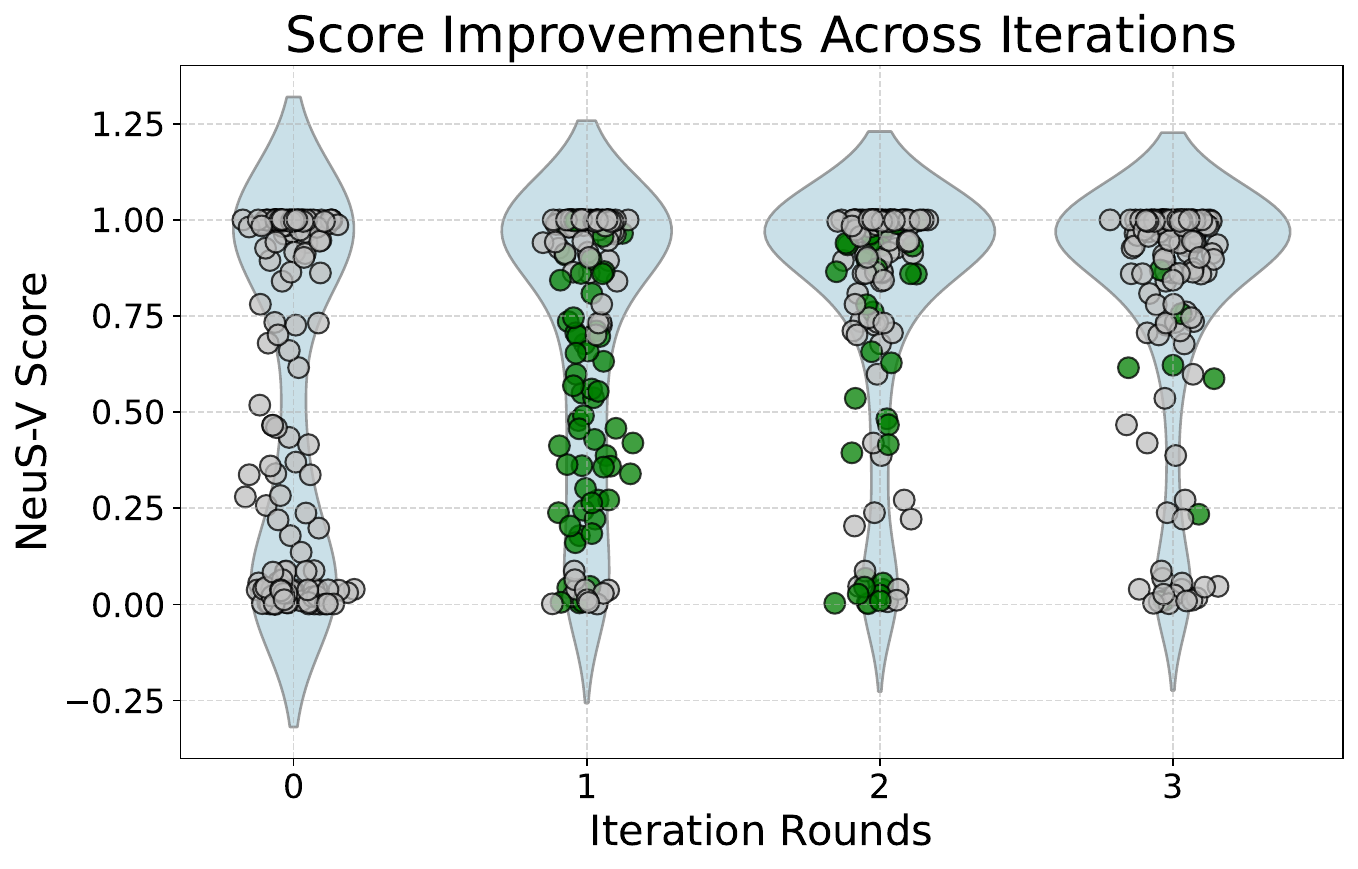}
                \caption{Pika-2.2}
                \label{fig:jitter-pika}
            \end{subfigure}
            \hfill
            \begin{subfigure}{0.32\linewidth}
                \centering
                \includegraphics[width=\linewidth]{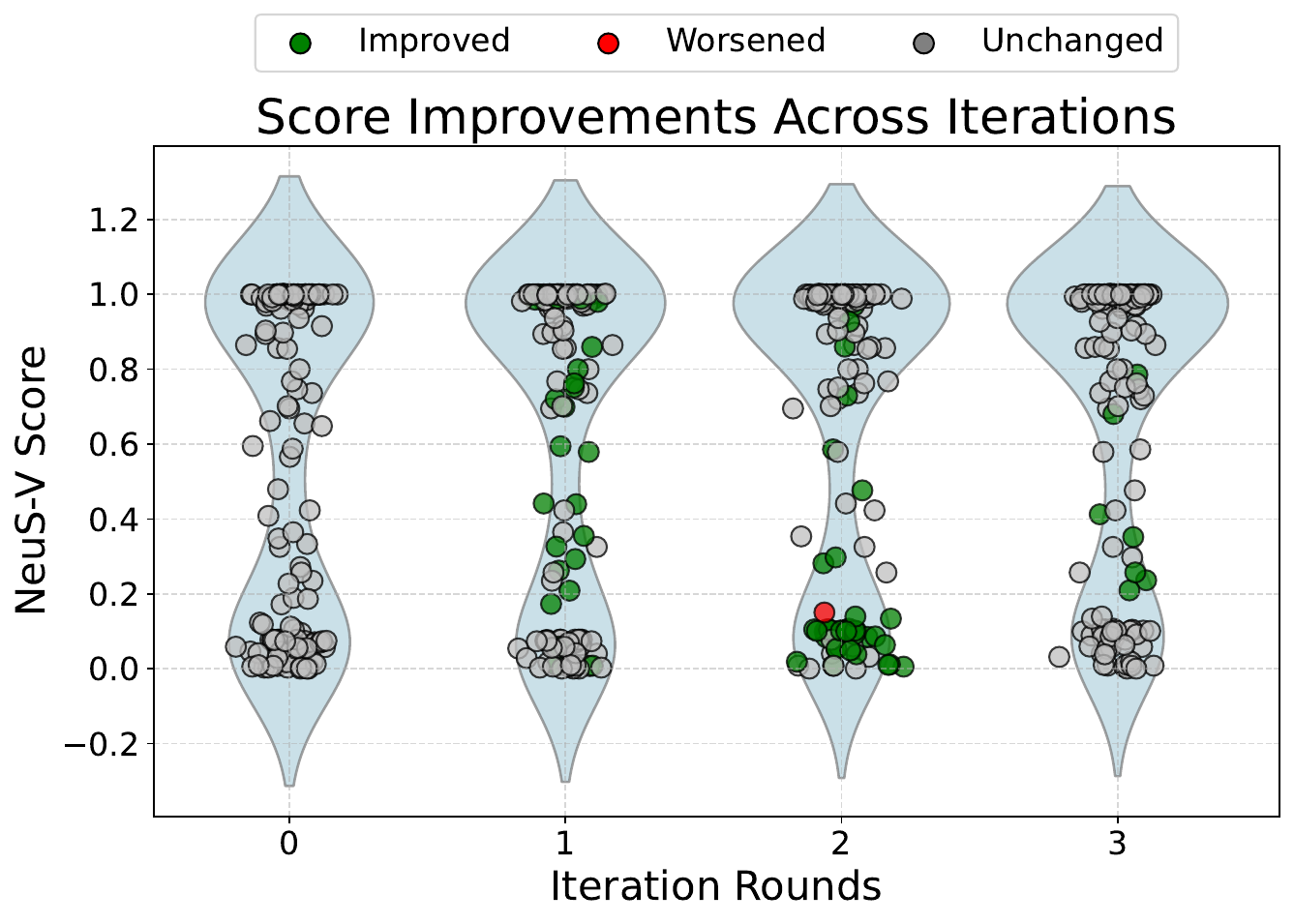}
                \caption{Gen-3}
                \label{fig:jitter-gen3}
            \end{subfigure}
            \hfill
            \begin{subfigure}{0.32\linewidth}
                \centering
                \includegraphics[width=\linewidth]{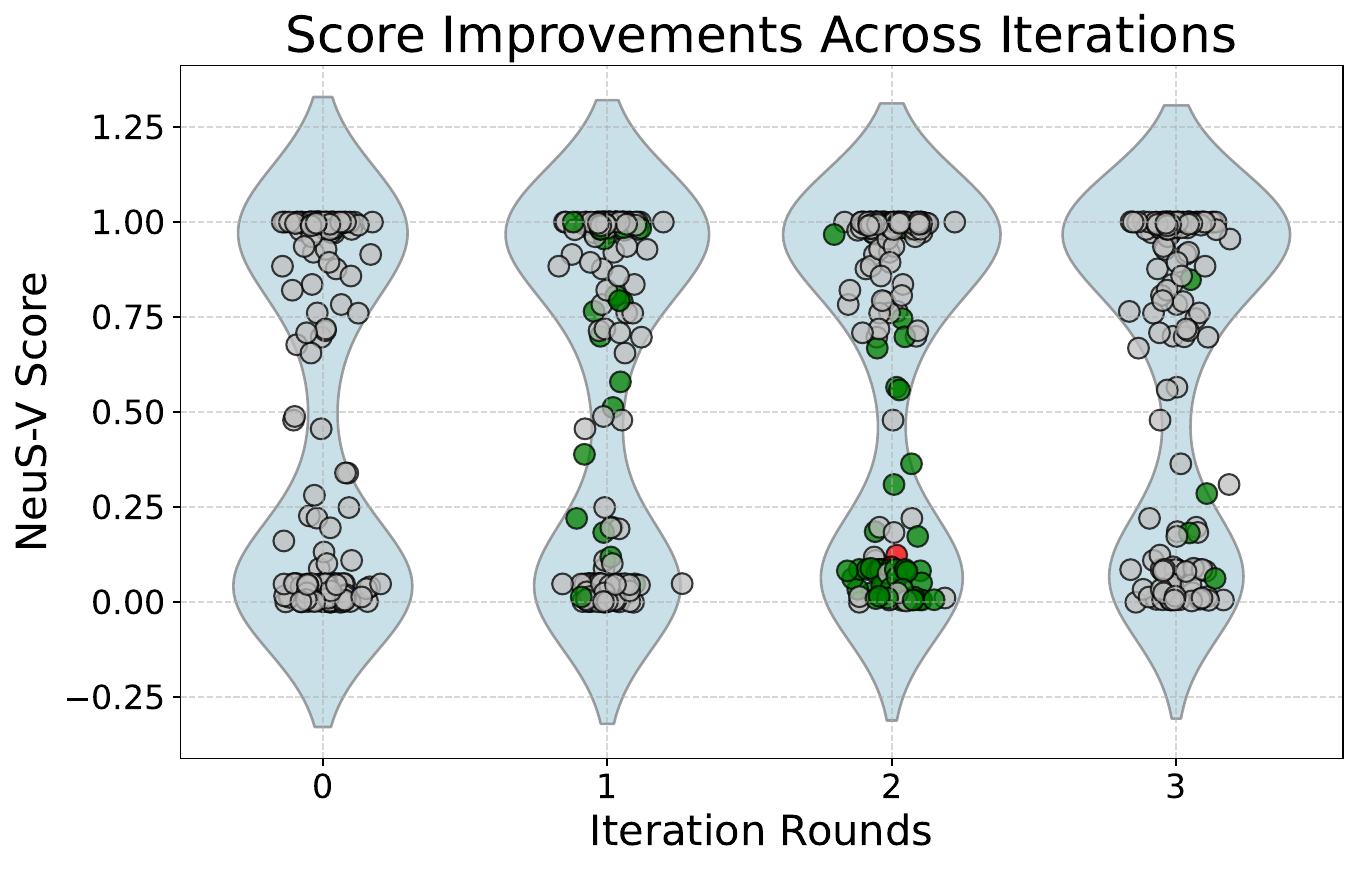}
                \caption{CogVideoX-5B}
                \label{fig:jitter-cogvideox}
            \end{subfigure}
            \caption{\textbf{Improvements from Iterative Rounds of Refinement.} Distribution of \neusv~score changes with a violin plot overlay. Green/red points indicate improvements/degradation per sample.}
            \label{fig:jitter-plot}
            \vspace{-2em}
        \end{figure}

    \subsection{Effect of Iterative Refinement}
        In our experiments, we perform three rounds of refinement, analyzing how each iteration contributes to improving temporal alignment. \Cref{fig:jitter-plot} visualizes these improvements. We observe that the violin plot widens progressively with each iteration, indicating increased variation in \neusv~scores after editing. Our results show that each round of refinement provides measurable improvements, but gains plateau around the third iteration. This suggests that while early rounds effectively correct temporal inconsistencies, further refinements yield diminishing returns. Additionally, we find that CogVideoX-5B responds poorly to \(\projectname\)’s edit instructions, leading to minimal improvement compared to other models, whereas Pika-2.2 responds well.


    \begin{wraptable}{r}{0.6\columnwidth}
        \centering
        \vspace{-4em}
        \caption{\textbf{Ablation Study on Key Frame Editing.} Comparison of \neusv~and VBench scores for Pika-2.2 with and without key frame editing. Since VBench scores are highly affected by key frame modifications, this is not our default architectural choice.}
        \resizebox{0.6\columnwidth}{!}{
            \begin{tabular}{ccccc}
                \toprule
                \multirow{2}{*}{Key Frame Editing} & \multicolumn{2}{c}{\neusv} & \multicolumn{2}{c}{VBench} \\
                \cmidrule(lr){2-3} \cmidrule(lr){4-5}
                & Original & Edited & Original & Edited \\
                \midrule
                \ding{55} & 0.577 & 0.811 \textcolor{teal}{(+0.233)} & 0.789 & 0.772 \textcolor{red}{($-$0.017)} \\
                \ding{51} & 0.577 & 0.835 \textcolor{teal}{(+0.258)} & 0.789 & 0.683 \textcolor{red}{($-$0.106)}  \\
                \bottomrule
            \end{tabular}
        }
        \vspace{-2em}
        \label{tab:ablation-keyframe}
    \end{wraptable}

\section{Discussion}
\label{sec:discussion}
\paragraph{On Scope and Positioning.}
    Our work does not aim to ``solve" T2V; instead, we pursue a narrow but high-impact goal: correcting the under-addressed failure mode of mis-ordered or missing events in today's T2V models, while tolerating a small, measured trade-off in aesthetic quality. Empirically, \projectname~improves temporal-logic recall by $+23.3$ points with only a $-1.7$ point drop on VBench. Our contribution goes beyond existing methods that assess only holistic temporal coherence \cite{sp_2025_CVPR}: \projectname~can pinpoint the weakest proposition in a temporal specification and localize its failure in time, revealing precisely \textit{where} and \textit{why} a video misaligns with the prompt. Moreover, because NeuS-E is a training-free, model-agnostic pipeline, it can be applied to \textit{any} black-box T2V generator through targeted segment-level edits, making it broadly practical in today's landscape dominated by closed-source and proprietary video models. 
    Finally, because \(\projectname\) regenerates only the short segment following a single identified keyframe
    which typically occurs near the middle or later portions of the video, this results in substantially fewer generated frames compared to restarting the generation process from the beginning. 
    This makes it substantially cheaper than whole-clip regeneration, step-by-step baselines, or the manual iterative rejection sampling currently employed by most users.

    \paragraph{On Evaluation Validity.}
    A natural concern is whether our evaluation is circular, since \neusv~informs 
    both the refinement signal and the primary metric. We emphasize that 
    \(\projectname\) does not optimize the \neusv~score directly; it uses only the 
    weakest-proposition trace for localization, not for score-based optimization. If 
    \neusv~merely rewarded re-generation, then full re-sampling or step-by-step 
    generation would yield similar gains---which our ablation in 
    \Cref{tab:ablation-generation-strategy} shows is not the case. Furthermore, 
    VBench scores remain stable after editing, and our T2VCompBench evaluation 
    (\Cref{tab:t2vcompbench}) provides fully independent confirmation, with an 
    average $+11\%$ improvement driven by gains in \texttt{Action} and 
    \texttt{Interaction}---the dimensions most closely tied to correcting missing 
    or misordered events.

    \paragraph{On the Necessity of Formal Verification over LVLMs.}
    One might ask whether a modern large vision-language model (LVLM) could 
    replace the neuro-symbolic pipeline for identifying misaligned segments. In 
    practice, direct LVLM querying produces coarse, sequence-level judgments 
    (\eg, ``the person never stands up''), but it does not yield the exact frame 
    index or transition point needed to determine where the edit should begin. 
    Our pipeline constructs a video automaton and applies model checking to 
    obtain a concrete execution trace, identifying the weakest proposition and 
    the precise frame at which it fails. This level of temporal resolution is 
    necessary for targeted refinement and cannot be produced by a single forward 
    pass of a standard LVLM. Moreover, most LVLMs process videos via sparse 
    frame sampling (\eg, 32--128 frames), introducing temporal aliasing: if the 
    critical misalignment occurs between sampled frames, the model cannot detect 
    the transition, let alone localize it.
    
\paragraph{Limitations.}
    While \projectname~consistently improves temporal fidelity, its effectiveness is constrained by the maturity of today's generation backbones. In particular, subject inconsistencies, flickering artifacts, implausible physics, and failures on abstract or highly stylized prompts expose the limits of current image and video generation models rather than our pipeline. Our ablation on keyframe editing with OmniGen (\Cref{tab:ablation-keyframe}) illustrates this trade-off: while incorporating keyframe edits further boosts \neusv~scores ($+0.258$), it also introduces a larger drop in VBench ($-0.106$), highlighting how \textit{aggressive edits can destabilize visual quality}. Hence, we default to the non-keyframe-editing variant, which strikes a better balance between temporal alignment and aesthetics.
    
    The key takeaway is that zero-training, neuro-symbolic feedback provides a principled and general mechanism for improving temporal alignment. As the underlying generation models advance, our approach will remain compatible and continue to enhance them, ensuring that temporal fidelity scales alongside improvements in visual quality.




\section{Conclusion}
\label{sec:conclusion}
    To conclude, we propose \(\projectname\), a training-free framework for improving text-to-video generation on temporally complex prompts. \(\projectname\) leverages neuro-symbolic feedback to diagnose semantic and temporal inconsistencies in generated videos and guide localized segment regeneration. Through this verification-guided refinement process, the system selectively corrects only the portions of the video responsible for specification violations while preserving segments that already satisfy the prompt. Empirical evaluations on established benchmarks and human preference studies demonstrate that \(\projectname\) consistently improves temporal fidelity across diverse themes and levels of prompt complexity. We hope this work encourages further exploration of neuro-symbolic approaches for advancing long-horizon video generation. 


%
%

\bibliographystyle{splncs04}
\bibliography{main}

\clearpage
\appendix

\section{Appendix: Methodology Clarifications}
\subsection{Temporal Logic Operation Example}
\label{sec:appx_temporal_logic}
Given a set of atomic propositions $\mathcal{P}=$ \{\text{Event A}, \text{Event B}\}, the TL specification $\Phi=$ \always~Event A (read as ``Always Event A") means that `Event A' is \texttt{True} for every step in the sequence. Additionally, $\Phi=$ \eventually~Event B (read as ``eventually event b") indicates that there exists at least one `Event B' in the sequence. Lastly, $\Phi=$ Event A \until \hspace{1pt} Event B (read as ``Event A Until Event B") means that `Event A' exists until `Event B' becomes \texttt{True}, and then `Event B' remains \texttt{True} for all future steps. 

\subsection{Text-to-Temporal-Logic}
We use the prompt below to decompose a text prompt into 1) a set of propositions and 2) a TL specification.

\vspace{1em}

\begin{minipage}{.95\linewidth}
    \begin{tcolorbox}[
        title=$f_{\text{T2TL}}$ Prompt for Text to TL Specification (.md),
        colback=white,
        colframe=gray,
        colbacktitle=gray
    ]
    \ttfamily\scriptsize
**System Message**:

\smallskip
Your input field is: \\
- `input\_prompt` (str): Input prompt summarizing what happened in a video. \\

\smallskip
Your output fields are: \\
- `input\_propositions` (str): A list of atomic propositions that correlate with the inputted prompt formatted as [proposition\_1, proposition\_2, ...].

- `output\_specification' (str): The formal specification of the inputted prompt. This is a temporal logic sequence made by combining the inputted propositions with temporal logic symbols.

\smallskip
Your objective is: \\
- Convert the prompt into a list of propositions and a temporal logic specification using the specified schema. \\
\smallskip
\hrule height 0.05pt
\medskip

**User message**:

\smallskip
\underline{Input Prompt}: A person meditates by the lake, and a few seconds later, stands up for a moment before leaving.

\smallskip
Respond with the corresponding output fields.

\bigskip
**Assistant message**:

\smallskip
\smallskip
\underline{Output Propositions}: [`person is meditating', `lake shore', `person is standing', `person is walking away']

\underline{Output Specification}: (person is meditating $\wedge$ lake shore) $\mathsf{X}$ person is standing $\mathsf{X}$ person is walking away
    \end{tcolorbox}
    \captionsetup{type=prompt}
    {\textbf{Prompt 1}: \textbf{Text to Specification Prompt.} System prompt to map prompts and propositions to the specification.}
    \label{prompt:t2p}
\end{minipage}

\subsection{Vision Language Model}
\label{sec:appx_vlm_calibration}
In this section, we provide the implementation details to detect the existence of propositions to label each frame in the synthetic video. We use the VLM to interpret semantics and extract confidence scores from $\mathcal{F}$ based on the text query $T$ . We pass each $\prop_i \in \propset$ along with the prompt to the VLM and calculate the token probability for the output response, which is either \texttt{True} or \texttt{False}. To calculate the token probability, we retrieve logits for the response tokens and compute the probability of that token after applying a softmax. Finally, the semantic confidence score is the product of these probabilities as follows:
        \begin{equation}
            \label{eq:vlm}
            c_i =\vlm(p_i, \mathcal{F}) \prod_{j=1}^{k} P\left( t_j \mid \prop_i, \mathcal{F}, t_1, \dots, t_{j-1} \right) \  \forall \, p_i \in \propset,
        \end{equation}
        where $(t_1,\dots,t_k)$ is the sequence of tokens in the response. Each term $P(t_j | \cdot) = \frac{e^{l_{j,t_j}}}{\sum_{z}e^{l_{j,z}}}$ is over the logits $l_{j,{t_j}}$ at position $j$, whereas $P(t_k | \cdot) = \frac{e^{l_{j,t_k}}}{\sum_{z}e^{l_{k,z}}}$ is over those at position $k$.
        
\subsubsection{Inference Via Vision Language Models}

We use a VLM as a semantic detector. We pass each atomic proposition $p_i \in \mathcal{P}$ such as ``person'', ``car'', ``person in the car'', etc. Once the VLM outputs either ``Yes'' or ``No'', we compute the token probability of the response and use it as a confidence score for the detection.

\subsubsection{False Positive Threshold Identification}
\paragraph{Dataset for Calibration:}
We utilize the COCO Captions \cite{chen2015microsoft} dataset to calibrate the following open-source vision language models -- InternVL2  Series (1B, 2B, 8B) \cite{internvl2024v2} and LLaMA-3.2 Vision Instruct \cite{metallama32v} -- for \neusv{}. Given that each image-caption pair in the dataset is positive coupling, we construct a set of negative image-caption pairs by randomly pairing an image with any other caption corresponding to a different image in the dataset. Once we construct the calibration dataset, which comprises 40000 image-caption pairs, we utilize the VLM to output a `Yes' or a `No' for each pair.

\vspace{10pt}
\begin{minipage}{0.95\linewidth}
    \begin{tcolorbox}[
        title=Prompt for Semantic Detector (VLM),
        colback=white,
        colframe=gray,
        colbacktitle=gray]
    \label{prompt:vlm}
        \ttfamily\scriptsize
        Is there \{atomic proposition ($\prop_i$)\} present in the sequence of frames?

        [PARSING RULE] 1. You must only return a Yes or No, and not both, to any question asked.
        
        2. You must not include any other symbols, information, text, or justification in your answer or repeat Yes or No multiple times.

        3. For example, if the question is 'Is there a cat present in the Image?', the answer must only be 'Yes' or 'No'.
    \end{tcolorbox}
    \captionsetup{type=prompt}
    {\textbf{Prompt 2}: \textbf{Semantic Detector VLM.} Used to identify the atomic proposition within the frame by initiating the VLM with a single frame or a series of frames.}
    \label{prompt:semantic-detector-vlm}
\end{minipage}

\paragraph{Thresholding Methodology}

\begin{figure}[t]
    \centering
    \includegraphics[width=0.7\linewidth, trim=1in 0 1in 0, clip]{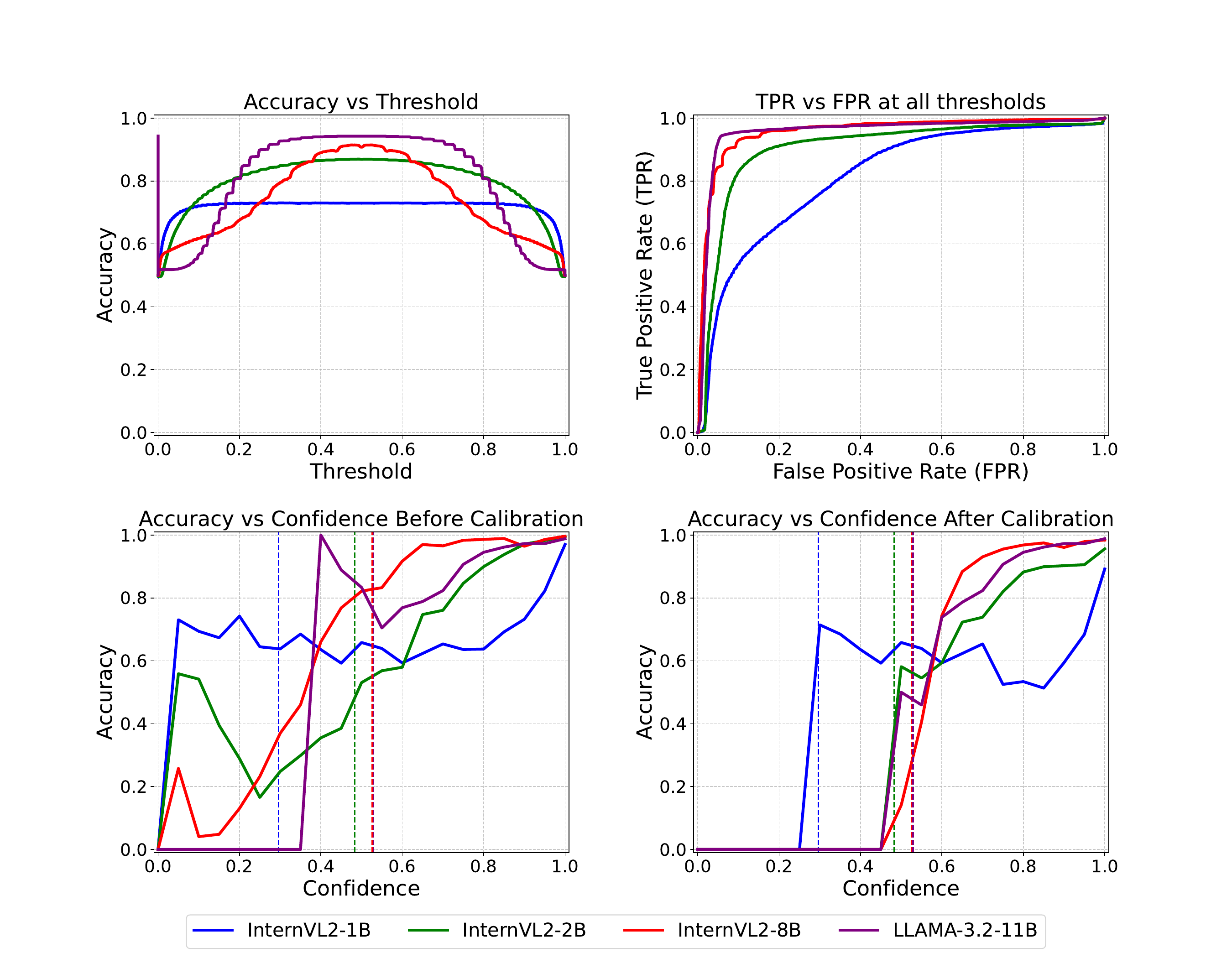}
    \caption{\textbf{Calibration Plots.} We plot the accuracy vs threshold for all VLMs on our calibration dataset constructed from the COCO Caption dataset (top left). We plot the True Positive Rate (TPR) vs False Positive Rate (FPR) across all thresholds on the top right. Finally, the bottom plots show the confidence vs accuracy of the model before and after calibration, respectively.}
    \label{fig:calibration}
\end{figure}

We can identify the optimal threshold for the VLM by treating the above problem as either a single-class or multi-class classification problem. We opt to do the latter. The process involves first compiling detections into a list of confidence scores and one-hot encoded ground truth labels. We then sweep through all available confidence scores to identify the optimal threshold. Here, we calculate the proportion of correct predictions by applying each threshold (see Figure \ref{fig:calibration}). The optimal threshold is identified by maximizing accuracy, which is the ratio of the true positive and true negative predictions. Additionally, to comprehensively evaluate model behavior, we compute Receiver Operating Characteristics (ROC), as shown in Figure \ref{fig:calibration}, by computing the true positive rate (TPR) and false positive rate (FPR) across all thresholds. Once we obtain the optimal threshold, we utilize it to calibrate the predictions from the VLM. We show the accuracy vs confidence plots before and after calibration in \Cref{fig:calibration}.

\subsection{Video Automaton Generation Function}
\label{sec:appx_va_construction}
Given a calibrated score set across all frames $\mathcal{F}_n$ (where $n$ is the frame index of the video) and propositions in $\propset$, we construct the video automaton $\va$ using the video automaton generation function (see \Cref{eq:va}).  
\begin{equation} 
    \label{eq:calibrated_score}
    \mathbb{C} = \{ \mathbb{C}_{p_i, j} \mid p_i \in \propset, j \in \{1, 2, \dots, n\} \}.
\end{equation}

As shown in \Cref{alg:av_generation}, we first initialize the components of the automaton, including the state set $Q$, the label set $\lambda$, and the transition probability set $\delta$, all with the initial state $q_0$. Next, we iterate over $\mathbb{C}$, incrementally constructing the video automaton by adding states and transitions for each frame. This process incorporates the proposition set and associated probabilities of all atomic propositions. We compute possible labels for each frame as binary combinations of $\propset$ and calculate their probabilities using the  $\mathbb{C}$.

\subsubsection{Edit Instruction Prompts}
We present different prompts to edit the keyframe and generate the new video segment in this section.
\vspace{2em}

\begin{minipage}{0.95\linewidth}
    \begin{tcolorbox}[
        title=Prompt for Image Editing ($T_{\text{img}}$),
        colback=white,
        colframe=gray,
        colbacktitle=gray]
    \label{prompt:vlm}
        \ttfamily\scriptsize
    Add \{Weak Proposition\} to the image
    \end{tcolorbox}
    \captionsetup{type=prompt}
    {\textbf{Prompt 3}: \textbf{Image (keyframe) Edit Instruction.} Used to edit the keyframe image to have missing semantics. The Weak Proposition will be given and passed along with the prompt}
\end{minipage}

\vspace{1em}
\begin{minipage}{0.95\linewidth}
    \begin{tcolorbox}[
        title=Prompt for New Video Generation ($T_{\text{video}}$),
        colback=white,
        colframe=gray,
        colbacktitle=gray]
    \label{prompt:vlm}
        \ttfamily\scriptsize
    You are tasked with refining video narratives generated by text-to-video models based on user feedback. For each case, you will receive two inputs:
   
    1. \textbf{Original Prompt}: A description of the intended video narrative.
    \\
    2. \textbf{Feedback}: Textual guidance on what is missing or needs adjustment in the video.
    \end{tcolorbox}
    \captionsetup{type=prompt}
    {\textbf{Prompt 4}: \textbf{New Video Segment Generation Instruction.} Used to generate a new prompt to generate a new video segment.}
\end{minipage}

\section{Appendix: Additional Results}

\subsection{VBench Scores}
We provide the VBench scores before and after editing. We see that the edited video shows very little degradation.

\begin{table}[h]
    \centering
    \caption{\textbf{VBench Scores Before and After Editing.} Comparison of original and edited VBench scores across different models.}
        \begin{tabular}{ccc}
            \toprule
            Model & Original & Edited \\
            \midrule
            Gen-3 & 0.789 & 0.772 \textcolor{red}{($-$0.017)} \\
            Pika-2.2 & 0.799 & 0.784 \textcolor{red}{($-$0.015)} \\
            CogVideoX-5B~\cite{yang2024cogvideox} & 0.672 & 0.660 \textcolor{red}{($-$0.012)} \\
            \bottomrule
        \end{tabular}
    \label{tab:vbench-original-edited}
\end{table}



\subsection{Human Study Annotation}
\begin{figure}[t]
    \centering
    \includegraphics[width=.4\linewidth]{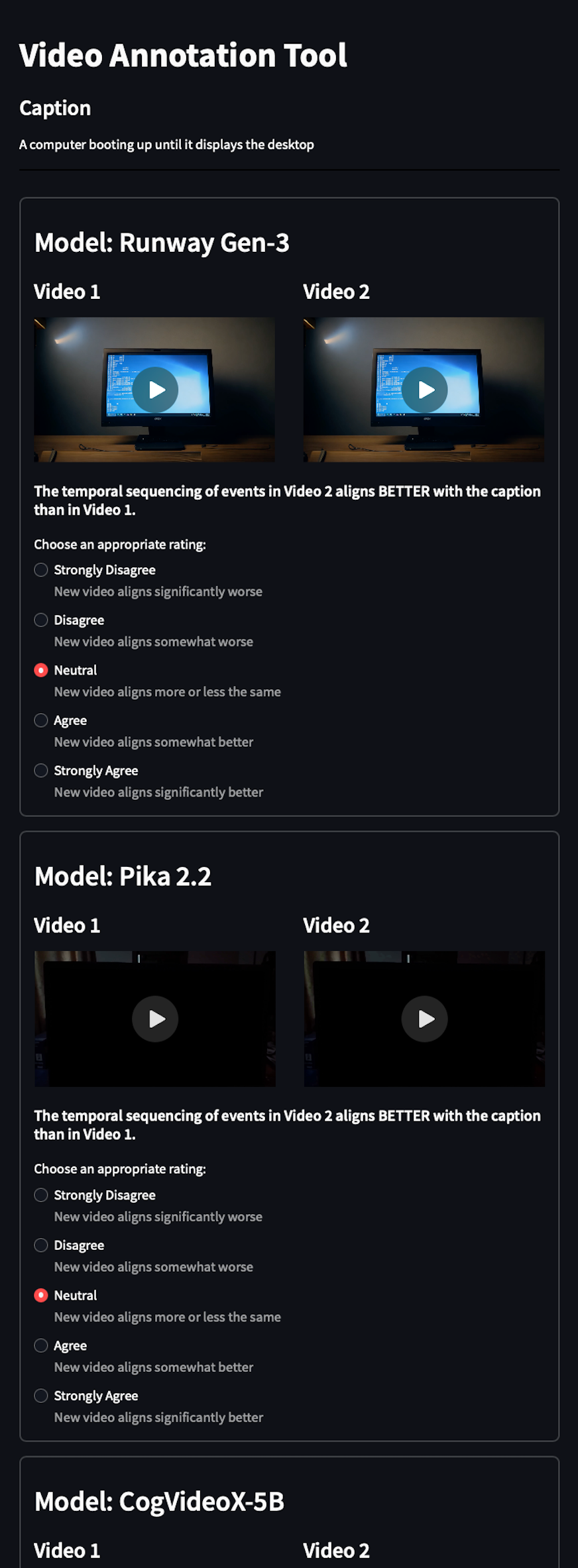}
    \caption{\textbf{Tool for Annotating Videos.} Subjects evaluate the video edited by \textit{NeuS-E} by comparing it to its original generation across five levels: strongly disagree, disagree, neutral, agree, and strongly agree.}
    \label{fig:annotation-tool}
\end{figure}

We provide the annotation tool (see \Cref{fig:annotation-tool}) for the randomized A/B testing of the temporal alignment of the generated videos with respect to the prompt. Here, Video 1 and Video 2 are randomly assigned as the original video and the edited videos for human judgment.

\section{Use of Large Language Models}

The author would like to acknowledge the use of Gemini, ChatGPT, and Claude in the preparation of this paper. The model served as a writing assistant, which was only used for providing valuable improvements to the clarity, conciseness, and logical flow of isolated sections, more specifically, the introduction, the methodology, and the results. It was also utilized, for assistance, with specific LaTeX formatting challenges. While the AI provided helpful refinement, the core scientific contributions, conceptual framework, and all final editorial decisions were the authors' own.

\section{Algorithms}
We present the algorithms for both \(\projectname\) and the video automaton generation in \Cref{alg:neus_e} and \Cref{alg:av_generation}.

\clearpage

\resizebox{0.75\linewidth}{!}{%
\begin{minipage}{\linewidth}
\begin{algorithm}[H]
    \footnotesize
    \DontPrintSemicolon
    \SetKwInOut{KwInput}{Input}
    \SetKwInOut{KwRequire}{Require}
    \SetKwInOut{KwOutput}{Output}
    \KwRequire{Text-to-video model \(\mathcal{M}_{\text{T2V}}\), prompt based image to image model \(\text{omnigen}\) Text-to-TL function \(f_{\text{T2TL}}(T)\), video automaton generation function \(\xi(\cdot)\), probabilistic model checking function \(\Psi(\cdot)\), edit instruction generation function \(\text{LLM}_{\text{EIG}}\), satisfaction probability threshold \(\theta\), maximum iterations \(\kappa\)}
    
    \KwInput{Text prompt \(T\), initial generated video \(\mathcal{V} = \mathcal{M}_{\text{T2V}}(T)\)}
    
    \KwOutput{Refined video \(\mathcal{V}^*\)}
    
    \Begin{
        \(\probsatis \leftarrow 0\) \;
        \(\textit{iter} \leftarrow 0\) \;
        \(\mathcal{V} \leftarrow \mathcal{M}_{\text{T2V}}(T)\) \tcp*{Generate a video} 
        \(\propset, \spec \leftarrow f_{\text{T2TL}}(T)\) \tcp*{Decompose a prompt} 
        
        \While{\(\mathbb{P}[\va \models \Phi] < \theta \land \text{iter} < \kappa\)}{
            \(\mathbb{C} \leftarrow \{\}\) \;
            
            \For{\(n = 0\) \KwTo \(\mathrm{length}(\mathcal{V})\)}{
                \For{\(\prop_i \in \propset\)}{
                    \(c_{i,n} \leftarrow \vlm(p_i, \mathcal{F}_n)\) \;
                }
                \(\mathbb{C} \leftarrow \mathbb{C} \cup \{c_{i,n}\}\) \;
            }
            \(\va \leftarrow \xi(\propset, \mathbb{C})\) \;
            \(\probsatis \leftarrow \Psi (\va, \spec)\) \;
            \If{\(\probsatis \geq \theta\)}{
                \Break \;
            }
            \tcp*{Identify the weakest proposition}
            \(\text{max\_delta} \leftarrow 0\) \;
            \(\wp \leftarrow \text{None}\) \;
            \(i^{*} \leftarrow \text{None}\) \;
            \For{\(\prop_i \in \propset\)}{
                \(\tilde{\mathbb{C}} \leftarrow \mathbb{C}\) \;
                \For{\(n = 0\) \KwTo \(\mathrm{length}(\mathcal{V})\)}{
                    \(\tilde{c}_{i,n} \leftarrow 1.0\) \;
                }
                \(\evai \leftarrow \xi(\propset, \tilde{\mathbb{C}})\) \;
                \(\eprobsatis \leftarrow \Psi(\evai, \spec)\) \;
                \(\delta_i \leftarrow \eprobsatis - \probsatis\) \;
                \If{\(\delta_i > \text{max\_delta}\)}{
                    \(\text{max\_delta} \leftarrow \delta_i\) \;
                    \(\wp \leftarrow \prop_i\) \;
                    \(i^{*} \leftarrow i\) \;
                }
            }
            \tcp*{Localize the influence of the weakest proposition}
            \(\text{max\_z} \leftarrow 0\) \;
            \(\frame^{*}_n \leftarrow \text{None}\) \;
            \(n^{*} \leftarrow \text{None}\) \;
            \For{\(n = 0\) \KwTo \(\mathrm{length}(\mathcal{V})\)}{
                \(\tilde{\mathbb{C}} \leftarrow \mathbb{C}\) \;
                \(\tilde{c}_{\wp,n} \leftarrow 1.0\) \;
                \(\text{evan} \leftarrow \xi(\propset, \tilde{\mathbb{C}})\) \;
                \(\enprobsatis \leftarrow \Psi(\evan, \spec)\) \;
                \If{\(\enprobsatis > \text{max\_z}\)}{
                    \(\text{max\_z} \leftarrow \enprobsatis\) \;
                    \(\frame^{*}_n \leftarrow \frame_n\) \;
                    \(n^{*} \leftarrow n\) \;
                }
            }
            \tcp*{Video refinement}
            \(\mathcal{V}_{\text{trimmed}} \leftarrow \mathcal{V}[0:n^{*}]\) \tcp*{Trim video up to impacted segment}
            \(T_{\text{new}} \leftarrow\text{LLM}_{\text{EIG}}(\wp),T_{\text{video}}\) \tcp*{Generate edit instruction for \(\wp\)} 
            \(\hat{\frame}^{*}_n \leftarrow  \text{omnigen}(\frame^{*}_n,T_{\text{img}})\) \tcp*{Edit keyframe based on the prompt} 
            \(\mathcal{V}_{\text{new}} \leftarrow \mathcal{M}_{\text{T2V}}(\hat{\frame}^{*}_n, T_{\text{new}})\) \tcp*{Generate new video segment} 
            \(\mathcal{V} \leftarrow \mathcal{V}_{\text{trimmed}} + \mathcal{V}_{\text{new}}\) \tcp*{Merge trimmed video with new segment with text prompt and keyframe} 
            \(\textit{iter} \leftarrow \textit{iter} + 1\) 
        }
    }
    \caption{\textit{NeuS-E}}
    \label{alg:neus_e}
\end{algorithm}
\end{minipage}%
}



\newpage
\resizebox{0.75\linewidth}{!}{%
\begin{minipage}{\linewidth}
    \begin{algorithm}[H]
    \DontPrintSemicolon
    \KwInput{Set of semantic score across all frames given all atomic propositions \{$\mathbb{C} = \mathbb{C}_{p_i, j} \mid p_i \in \propset, j \in \{1, 2, \dots, n\}$\}, set of atomic propositions $\propset$}
    \KwOutput{Video automaton $\va$}
    \Begin{
        $Q \leftarrow \{q_0\}$ \tcp*{Initialize the set of states with the initial state}
        
        $\lambda \leftarrow \{(q_0, \text{initial})\}$ \tcp*{Initialize the set of labels with the initial label}
        
        $\delta \leftarrow \{\}$ \tcp*{Initialize the set of state transitions}
        
        $Q_p \leftarrow \{q_0\}$ \tcp*{Track the set of previously visited states}
        
        $n \leftarrow \frac{|\mathbb{C}|}{|\propset|}$ \tcp*{Calculate the total number of frames $n$}
        
        \For{$j \gets 1$ \KwTo $n$}{
            $Q_c \leftarrow \{\}$ \tcp*{Track the set of current states}
            
            \For{$e_j^k \in 2^{|\propset|}$}{
                \tcp{\( e_j^k \): unique combination of 0s and 1s for atomic propositions in \( \propset \)}
                $\lambda(q_j^k) = \{v_1, v_2, \dots, v_i \mid v_i \in \{1,0\}, \forall i \in \{1, 2, \dots, |\propset|\}\}$ 

                $pr(j, k) \leftarrow 1$ \tcp*{Initialize probability for the label}
                
                \For{$v_i \in \lambda(q_j^k)$}{
                    \tcp{Calculate probability for $e_j^k$}
                    \If{$v_i = 1$}{
                        $pr(j, k) \leftarrow pr(j, k) \cdot \mathbb{C}_{p_i, j}$
                    }
                    \Else{
                        $pr(j, k) \leftarrow pr(j, k) \cdot (1 - \mathbb{C}_{p_i, j})$
                    }
                
                \tcp{Add state and define transitions if the probability is positive}
                \If{$pr(j, k) > 0$}{
                    $Q \leftarrow Q \cup \{q_{j}^k\}$
    
                    $Q_c \leftarrow Q_c \cup \{q_{j}^k\}$
                    
                    $\lambda \leftarrow \lambda \cup \{(q_{j}^k, \lambda(q_j^k))\}$
        
                    \For{$q_{j-1} \in Q_p$}{
                    $\delta(q_{j-1}, q_j^k) \leftarrow pr(j, k)$
                    
                    $\delta \leftarrow \delta \cup \{\delta(q_{j-1}, q_j^k)\}$}
                    
                    \EndFor
                    }
                    
                \EndFor
                }
            $Q_p \leftarrow Q_c$ \tcp*{Update previous state}
            
            \EndFor    
            }
        
        \EndFor
        }    
        \tcp{Add terminal state}
        $Q \leftarrow Q \cup \{q_{j+1}^0\}$
        
        $\lambda \leftarrow \lambda \cup \{(q_{j+1}^0, \text{terminal})\}$
    
        \For{$q_{j-1} \in Q_p$}{
        $\delta(q_{j-1}, q_{j+1}^0) \leftarrow 1$
        
        $\delta \leftarrow \delta \cup \{\delta(q_{j-1}, q_{j+1}^0)\}$}
        \EndFor
    \tcp{Return video automaton}
    $\va \leftarrow (Q, q_0, \delta, \lambda)$ \\
    \Return $\va$\;
    }
    \caption{Video Automaton Generation}
    \label{alg:av_generation}
    \end{algorithm}
\end{minipage}%
}

\end{document}

%% file: preamble.tex
\usepackage{lipsum} 
\usepackage{pifont} 
\usepackage{tcolorbox}
\usepackage{amsmath}
\usepackage{amsfonts}
\usepackage{subcaption}
\usepackage{wrapfig}
\usepackage{ulem}
\usepackage{enumitem} 
\usepackage{quoting} 

\usepackage{float}
\usepackage[linesnumbered, ruled, vlined]{algorithm2e}
\newcommand{\algorithmstyle}[1]{\renewcommand{\algocf@style}{#1}}

\SetCommentSty{mycommfont}
\SetKwInput{KwInput}{Input}                
\SetKwInput{KwOutput}{Output}              
\SetKwInput{KwPrompt}{Prompts}             
\SetKw{KwStep}{step}
\SetKw{EndFor}{end for}
\SetKw{Require}{Require}
\SetKw{Break}{Break}

\usepackage{ifthen} 
\usepackage{etoolbox} 
\newboolean{showcomments}
\setboolean{showcomments}{true} 
\newrobustcmd{\sps}[1]{\ifthenelse{\boolean{showcomments}}{\textcolor{orange}{(\textbf{SP}: #1)}}{}}
\newrobustcmd{\mk}[1]{\ifthenelse{\boolean{showcomments}}{\textcolor{purple}{(\textbf{Minkyu}: #1)}}{}}
\newrobustcmd{\sas}[1]{\ifthenelse{\boolean{showcomments}}{\textcolor{red}{(\textbf{Sahil}: #1)}}{}}
\newrobustcmd{\merc}[1]{\ifthenelse{\boolean{showcomments}}{\textcolor{blue}{(\textbf{HG}: #1)}}{}}
\usepackage[hang,flushmargin]{footmisc}
\usepackage{multirow} 


%% file: macros.tex
\newcommand{\propset}{\mathcal{P}}
\newcommand{\prop}{p}
\newcommand{\ep}{\tilde{p}_i}
\renewcommand{\wp}{p^{*}_i}

\newcommand{\spec}{\Phi}

\newcommand{\always}{$\Box$}
\newcommand{\eventually}{$\diamondsuit$}
\newcommand{\ournext}{$\mathsf{X}$}
\newcommand{\until}{$\mathsf{U}$}

\newcommand{\va}{\mathcal{A}_{\mathcal{V}}}
\newcommand{\evai}{\tilde{\mathcal{A}}_{\mathcal{V},i}}
\newcommand{\evan}{\tilde{\mathcal{A}}_{\mathcal{V},n}}

\newcommand{\probsatis}{\mathbb{P}[\va \models \Phi]}
\newcommand{\eprobsatis}{\mathbb{P}[\evai \models \Phi]}
\newcommand{\enprobsatis}{\mathbb{P}[\evan \models \Phi]}
\newcommand{\neusv}{\textit{NeuS-V}}
\newcommand{\projectname}{\textit{NeuS-E}}

\newcommand{\video}{\mathcal{V}}
\renewcommand{\frame}{\mathcal{F}}

\newcommand{\vlm}{\mathcal{M}_{\text{VLM}}}